\documentclass[letterpaper, 10 pt, conference]{ieeeconf}  

\IEEEoverridecommandlockouts                              

\usepackage[dvipdfmx]{graphicx}
\let\labelindent\relax
\usepackage{enumitem}
\usepackage{bm}
\usepackage{amsmath,amssymb}
\usepackage{mathtools}
\usepackage{physics}
\usepackage{listings}
\usepackage{url}
\usepackage{subfigmat}
\usepackage{color}
\usepackage{multirow} 
\usepackage{ulem}

\newcommand{\doubleunderline}[1]{\underline{\underline{#1}}}

\title{\LARGE \bf
Goal Estimation-based Adaptive Shared Control\\ for Brain-Machine Interfaces Remote Robot Navigation
}

\author{Tomoka Muraoka$^{1}$ Tatsuya Aoki$^{1}$ Masayuki Hirata$^{2}$ Tadahiro Taniguchi$^{3}$ Takato Horii$^{1}$ and Takayuki Nagai$^{1}$
\thanks{*This work was supported by the Japan Science and Technology Agency (JST) Moonshot R\&D Grant Number JPMJMS2011.}
\thanks{$^{1}$ Dept. of Systems Innovation, Graduate School of Engineering Science, Osaka University, Osaka, Japan} 
\thanks{\tt\small \{t.muraoka@rlg., t.aoki@rlg., takato@, nagai@\} sys.es.osaka-u.ac.jp}
\thanks{$^{2}$ Dept. of Neurological Diagnosis and Restoration, Graduate School of Medicine, Osaka University, Osaka, Japan} 
\thanks{\tt\small mhirata@ndr.med.osaka-u.ac.jp}
\thanks{$^{3}$ Dept. of Science and Engineering, Ritumeikan University, Shiga, Japan}
\thanks{\tt\small taniguchi@em.ci.ritsumei.ac.jp}
}
\begin{document}

\maketitle
\thispagestyle{empty}
\pagestyle{empty}

\begin{abstract}
In this study, we propose a shared control method for teleoperated mobile robots using brain-machine interfaces (BMI). 
The control commands generated through BMI for robot operation face issues of low input frequency, discreteness, and uncertainty due to noise.
To address these challenges, our method estimates the user's intended goal from their commands 
and uses this goal to generate auxiliary commands through the autonomous system that are both at a higher input frequency and more continuous.
Furthermore, by defining the confidence level of the estimation, we adaptively calculated the weights for combining user and autonomous commands, thus achieving shared control.
We conducted navigation experiments in both simulated environments and participant experiments in real environments including user ratings, using a pseudo-BMI setup. 
As a result, 
the proposed method significantly reduced obstacle collisions in all experiments. 
It markedly shortened path lengths under almost all conditions in simulations 
and, in participant experiments, especially when user inputs become more discrete and noisy (p$<$0.01). 
Furthermore, under such challenging conditions, it was demonstrated that 
users could operate more easily, with greater confidence, and at a comfortable pace through this system.
\end{abstract}

\section{INTRODUCTION}
The potential of brain-machine interfaces (BMI) to enable remote control of robots offers significant opportunities for enhancing social participation among individuals with physical disabilities. 
This is achieved by providing essential navigation capabilities. 
However, the commands generated by BMI face unique challenges \cite{tonin2021noninvasive}.
Firstly, BMI systems often operate at low input frequencies, which can delay the control responsiveness. 
Secondly, the commands are discrete, limiting users to a set of predefined actions, such as moving forward, backward, or turning. 
Finally, noise interference poses a significant risk of inaccuracies in interpreting brain waves, leading to the potential issuance of incorrect commands to the robot \cite{chaudhary2022, mitchell2023assessment, yanagisawa2012electrocorticographic, vansteensel2016}.


Shared controls (SCs) are used to address these issues.
Kong et al.\cite{Kong20} and Tonin et al.\cite{Tonin20} proposed methods for navigating mobile robots using SCs to avoid obstacles. 
However, these approaches limit the application of autonomous assistance to obstacle avoidance.
In contrast, Zeng et al.\cite{Zeng20} proposed a method that dynamically combined user-input commands with commands autonomously generated from the system's autonomy for tasks involving moving the end effector of a robotic arm in a two-dimensional plane from a starting position to a target object.
In their method, users specify the target object they wish to grasp at the start of the task using gaze input. 
Subsequently, using gaze input and BMI, users input velocity commands to the gripper of the robot arm ($\bm{a}^{user}$). 
Simultaneously, the system autonomously generates velocity commands to move towards the target or avoid obstacles ($\bm{a}^{auto}$). 
These user and autonomous commands are combined using Equation (\ref{eq:shared_control}) and input into the robotic arm.
\begin{equation}
\label{eq:shared_control}
    \bm{a}^{shared} = (1-\alpha) \cdot \bm{a}^{user} + \alpha \cdot \bm{a}^{auto}
\end{equation}
This method of blending two commands is a widely used SC teleoperation paradigm (\cite{Carlson12, Gopinath17, Muelling2015}).
How to determine $\bm{a}^{auto}$ and the weight $\alpha$ is a central question in SC \cite{dragan2013}.

As a method to determine $\bm{a}^{auto}$,
Zeng et al.\cite{Zeng20} calculated it to direct towards the target object specified by the user at the start of a task while avoiding static obstacles.
However, this approach requires users to specify the target object explicitly at the beginning of the task, which presents issues, such as the need to re-specify the target upon changing it and the inability to accommodate targets that assume continuous values.
%
Conversely, some approaches determine $\bm{a}^{auto}$ by estimating the user's objective. 
Shervin et al.\cite{Shervin18} and Gottardi et al.\cite{Gottardi22} 
prepared a set of $N$ graspable target candidates 
for robot arm reaching tasks, 
and using the user's actions, they determined $\bm{a}^{auto}$ to direct towards the target with the highest probability among those candidates.
However, their method requires a predefined set of candidate goals.
Additionally, as it can only represent discrete goals, it is insufficient for application in navigation tasks.
Conversely, Beraldo et al.\cite{Beraldo22} proposed a method for determining $\bm{a}^{auto}$ in navigation tasks by determining a robot's intermediate destinations or "subgoals." 
The subgoals are calculated using a probability map that combines user commands with information from the robot's sensors. 
The velocity that follows the trajectory towards these subgoals becomes $\bm{a}^{auto}$. 
Their approach utilized probability maps to consider more continuous positions as potential goal candidates. 
They adopted short-term goals or subgoals to achieve SC that does not rely on maps.

Accordingly, we propose a method that uses user inputs to estimate the user's intended goal as a probability distribution, and we then calculate $\bm{a}^{auto}$ using the probability distribution.
Our proposed method computed the probability for all possible positions that the robot can assume and achieve in real time by accelerating the calculations with neural networks (U-Net). 
Furthermore, considering that static obstacles are often present when navigating with a robot, and obtaining maps is cost-effective, we focused on using maps to estimate long-term goals.

Furthermore, in SC, the weight $\alpha$, which combines user and autonomous system commands, plays a crucial role.
However, determining how to adaptively determine the weight parameter remains a long-standing issue.
As aforementioned, Beraldo et al.\cite{Beraldo22} treated user commands and environmental information equally to determine subgoals. 
Consequently, the weight reflecting the influence of user commands is fixed and does not vary. 
However, this weight should change adaptively depending on the environment and other factors.
In contrast, there are studies that dynamically adjust weights according to the environment \cite{Zeng20, Baoguo23}. 
Zeng et al.\cite{Zeng20} determined it by considering the distance between the target object speciﬁed by the user at the beginning of the task and end effector, as well as the distance between the end effector and obstacles.
However, this method encounters issues due to the necessity of predetermining the target.

Such a parameter $\alpha$ should reflect the robot's confidence. 
To calculate such confidence, the robot should be aware of the correctness of its behavior in relation to its goal. 
In this study, we utilized the estimated goal to calculate this confidence. 
Our method allows the simultaneous determination of this weight $\alpha$ once the goal has been estimated.

The main contributions of this paper are as follows: 
\begin{enumerate}
    \item To realize adaptive SC we proposed 
    goal estimation, 
    which minimizes the impact of noise-induced uncertainty using a fixed number of user command histories. 
    In addition, by employing a neural network (NN) with a U-Net architecture, the goal probabilities for all positions reachable by the robot can be rapidly calculated.
    Using the estimated goals, we generated autonomous commands that were continuous and at a high input frequency.
    
    \item 
    By defining the confidence level of the estimation, $c$, 
    we adaptively calculated the weight $\alpha$ for combining user and autonomous commands, thus realizing SC.
    
    \item 
    We demonstrated that the system integrating 1 and 2 significantly improves the performance of pseudo-BMI shared navigation tasks. 
\end{enumerate}

\section{PROPOSED METHOD}

\subsection{Overall View of the Proposed System}
\begin{figure*}[t]
    \centering
    \includegraphics[keepaspectratio, scale=0.5]{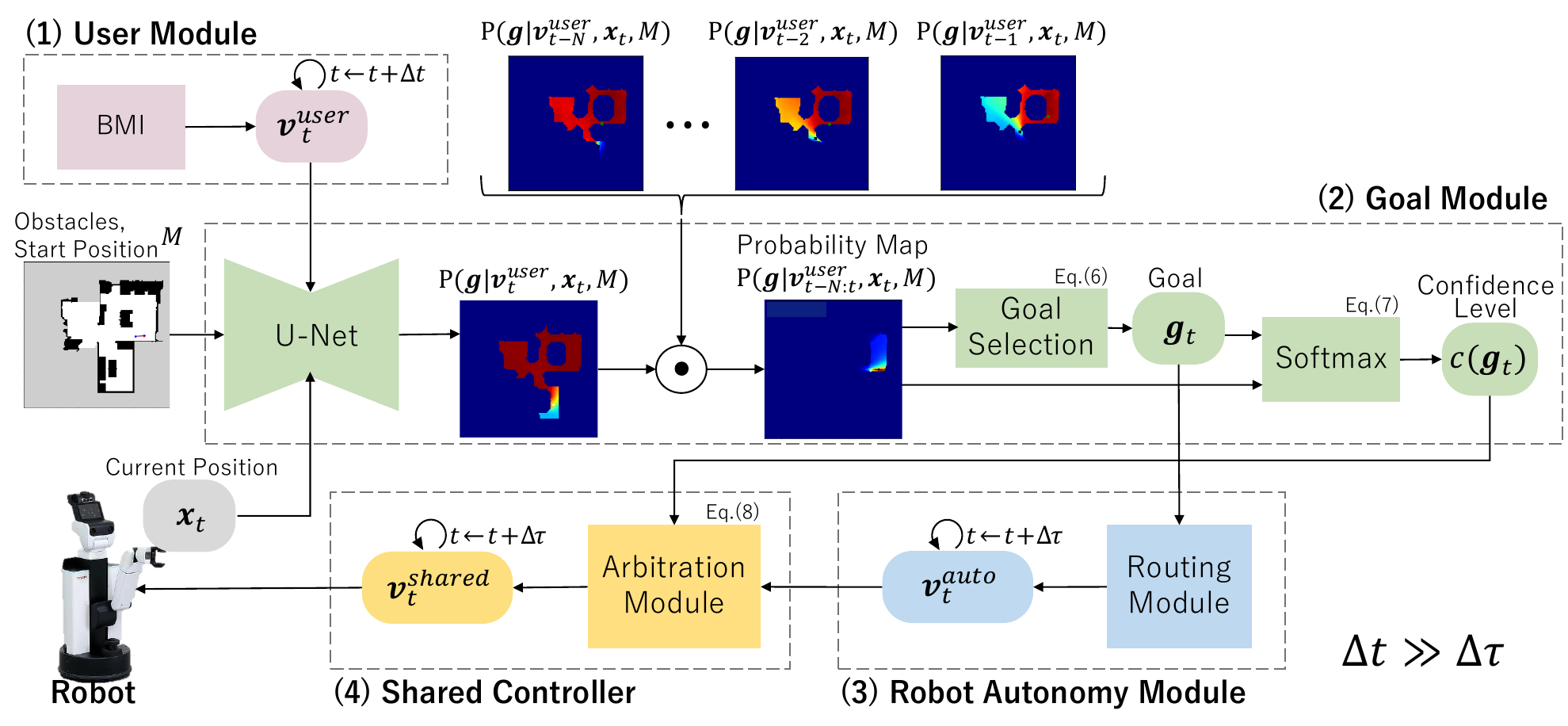}
    \caption{Overview of the proposed system:
    (1) The user module receives inputs from the user and defines them as velocity commands $\bm{v}^{user}_t$ for the robot, which are then output to the other modules. 
    (2) The goal module estimates the user's intended goal $bm{g}_t$ using environmental information and the user command, as well as calculates the confidence level $c(\bm{g}_t)$ of this estimation. 
    (3) The robot autonomy module generates autonomous commands $\bm{v}^{auto}_t$ to navigate towards the goal estimated by the goal module. 
    (4) The shared controller combines the user command and the autonomous command based on the confidence level of the estimation calculated by the goal module to produce the shared control command $\bm{v}^{shared}_t$, which dictates the robot's movement.}
    \label{fig:overview}
\end{figure*}

The SC system proposed in this study primarily comprises four modules, as shown in Fig.\ref{fig:overview}. 
This system was implemented using the robot operating system (ROS) \footnote{\url{https://wiki.ros.org/Documentation}}. Additionally, it was assumed in this study that the map was pre-learned.
\begin{enumerate} 
\renewcommand{\labelenumi}{(\arabic{enumi})}
    \item {\bf User Module}\\
    This module interprets the inputs from the user and translates them into velocity commands for the robot. 
    These commands, termed "user commands," are characterized by challenges, such as low input frequency, discreteness, and uncertainty due to noise.
    
    \item {\bf Goal Module}\\
    This module is pre-informed of the map and the robot's initial position, and continuously receives the robot's current location. Upon receiving a user command from (1), the goal module estimates the probability distribution of the user’s intended goal. Utilizing a fixed number of user-command histories in this calculation reduces the impact of noise-induced uncertainty.
    Subsequently, this module selects a single point as the user's intended goal using this probability distribution and calculates the confidence level of this estimation.
    
    \item {\bf Robot Autonomy Module}\\
    This module independently generates a route to the goal identified by (2) and calculates the necessary velocity at the current position to follow this route. 
    The velocity command generated by this process is referred to as an "autonomous command." 
    These autonomous commands, in comparison to user commands, have a sufficiently high input frequency and are continuous.
    In this study, path planning and velocity calculations were conducted using move base package \footnote{\url{https://wiki.ros.org/move_base}}, a standard package for ROS.
    
    \item {\bf Shared Controller}\\
    This receives the user command from (1) and the autonomous command from (3), 
    integrating them based on the confidence level determined by (2). 
    The outcome is a new velocity command, called "shared control command," which directs the movement of the robot. 
    The commands, similar to autonomous commands, also have a high input frequency and are continuous.
\end{enumerate}

\subsection{Goal Estimation}
The estimation of the user's intended goal in the goal module is performed in two steps.
First, using the user command $\bm{v}^{user}_t$ at time $t$, 
the probability distribution $P(\bm{g} \mid \bm{v}_{t-N:t}, \bm{x}_t, M)$ is calculated. 
This distribution represents the likelihood that the user aims for each position on the map. 
Subsequently, using that probability distribution, a goal $\bm{g}_t$ is selected, and the confidence level of that estimated goal $c(\bm{g}_t)$ is calculated.



\subsubsection{Definition of the Probability Distribution}
When the user command $\bm{v}^{user}_t$ is given, the probability distribution of the goal location intended by the user is defined as follows:
\begin{equation}
\label{probability_T}
    P(\bm{g} \mid \bm{v}^{user}_{t-N:t}, \bm{x}_t, M) =
    \frac{\prod_{t=t-N}^{t} P(\bm{g} \mid \bm{v}^{user}_{t}, \bm{x}_t, M)}{\sum_{\bm{g}}  \prod_{t=t-N}^{t} P(\bm{g} \mid \bm{v}^{user}_{t}, \bm{x}_t, M)}
\end{equation}
Here, $\bm{x}_t=(x_t, y_t)$ denotes the robot's position at time $t$, and $M$ represents the map indicating static obstacles.
In addition, $N$ represents the parameter that determines the number of steps back into the user's input history. 
By choosing $N$ sufficiently large, it is possible to ignore the influence of commands miscommunicated due to errors in the decoding of brain waves.
\begin{math} P(\bm{g} \mid \bm{v}^{user}_{t}, \bm{x}_t, M) \end{math} is defined as: 
\begin{equation}
\begin{split}
\label{probability_t}
&P(\bm{g} \mid \bm{v}^{user}_{t}, \bm{x}_t, M) \\ 
&= 
\exp{
-\sqrt{\qty(\frac{ \partial \phi(\bm{x}=\bm{x}_t, \bm{g}, M)}{\partial \bm{x}_t}  - \bm{v}^{user}_t )^2}
}
\end{split}
\end{equation}
%
Here, $\phi(\bm{x},\bm{g}, M)$ denotes a potential field when position $\bm{g}$ is considered the goal. 
The potential field represents a spatial distribution of potential values used to influence the movement of a robot, where lower values attract and higher values repel, guiding the robot towards the goal while avoiding obstacles.
In this study, we adopt the algorithm used in navfn\footnote{\url{https://wiki.ros.org/navfn}}, one of the global planner plugins provided by ROS, for defining the potential field.

\subsubsection{Calculating \begin{math} P(\bm{g} \mid \bm{v}^{user}_{t}, \bm{x}_t, M) \end{math} Using a NN}
\label{section:model_architecture}
\begin{figure*}[t]
    \centering
    \includegraphics[keepaspectratio, scale=0.5]{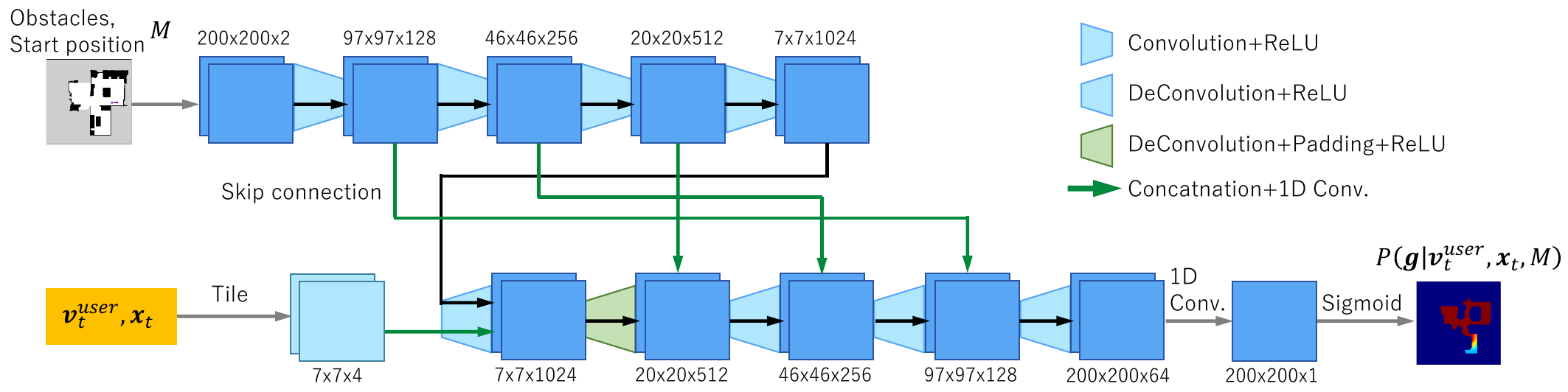}
    \caption{Neural network architecture. 
    The network takes as input 
    the map $M$, 
    the robot's initial position $\bm{x}_0$, 
    the robot's self-position $\bm{x}_t$ at time $t$, 
    and the user command $\bm{v}^{user}_t$ at time $t$, 
    and outputs \begin{math} P(\bm{g} \mid \bm{v}^{user}_{t}, \bm{x}_t, M) \end{math}, given the user command at time $t$.}
    \label{fig:model_architecture}
\end{figure*}

In the proposed method, we calculated the probability of a location being the user's goal for all positions reachable by the robot. 
Thus, it is necessary to compute Equation (\ref{probability_t}) for each point on the map. 
Performing this calculation every time a user command is received is time-consuming and can impair the system's real-time performance.
Therefore, we used a NN to compute Equation (\ref{probability_t}).
The architecture of the NN used in this study is shown in Fig. \ref{fig:model_architecture}, 
and is based on the U-Net structure.
The input to this NN consisted of a map $M$ (Fig.\ref{fig:training_data}(b)), 
the robot's initial position $\bm{x}_0$, 
its position at time $t$, $\bm{x}_t$, 
and the user command at time $t$, $\bm{v}^{user}_t$. 
The output is the calculation result of \begin{math} P(\bm{g} \mid \bm{v}^{user}_{t}, \bm{x}_t, M) \end{math} from Equation (\ref{probability_t}) when $\bm{v}^{user}_t$ is inputted (Fig. \ref{fig:training_data}(c)). 
\subsubsection{How to Train the NN}
\begin{figure}[t]
    \begin{subfigmatrix}{3}
        \subfigure[Map created using SLAM.]{\includegraphics[width=0.3\linewidth]{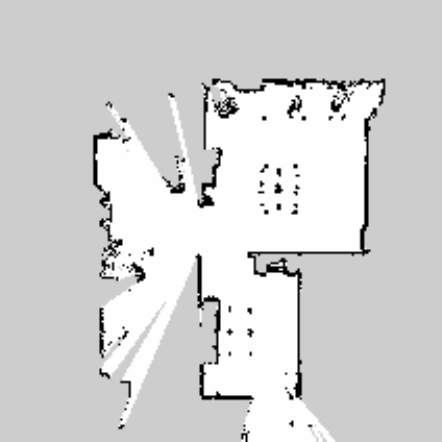}}
        \hspace*{-0.3cm}
        \subfigure[Map inputted into the model.]{\includegraphics[width=0.3\linewidth]{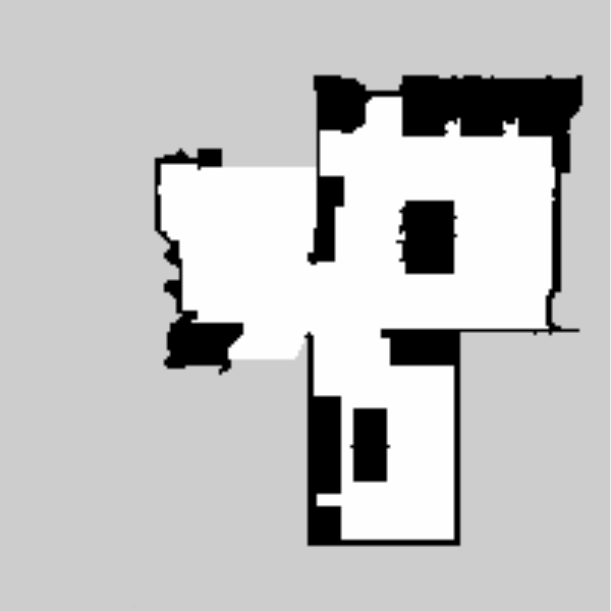}}
        \hspace*{-0.5cm}
        \subfigure[Example of an output of the model.]{\includegraphics[width=0.37\linewidth]{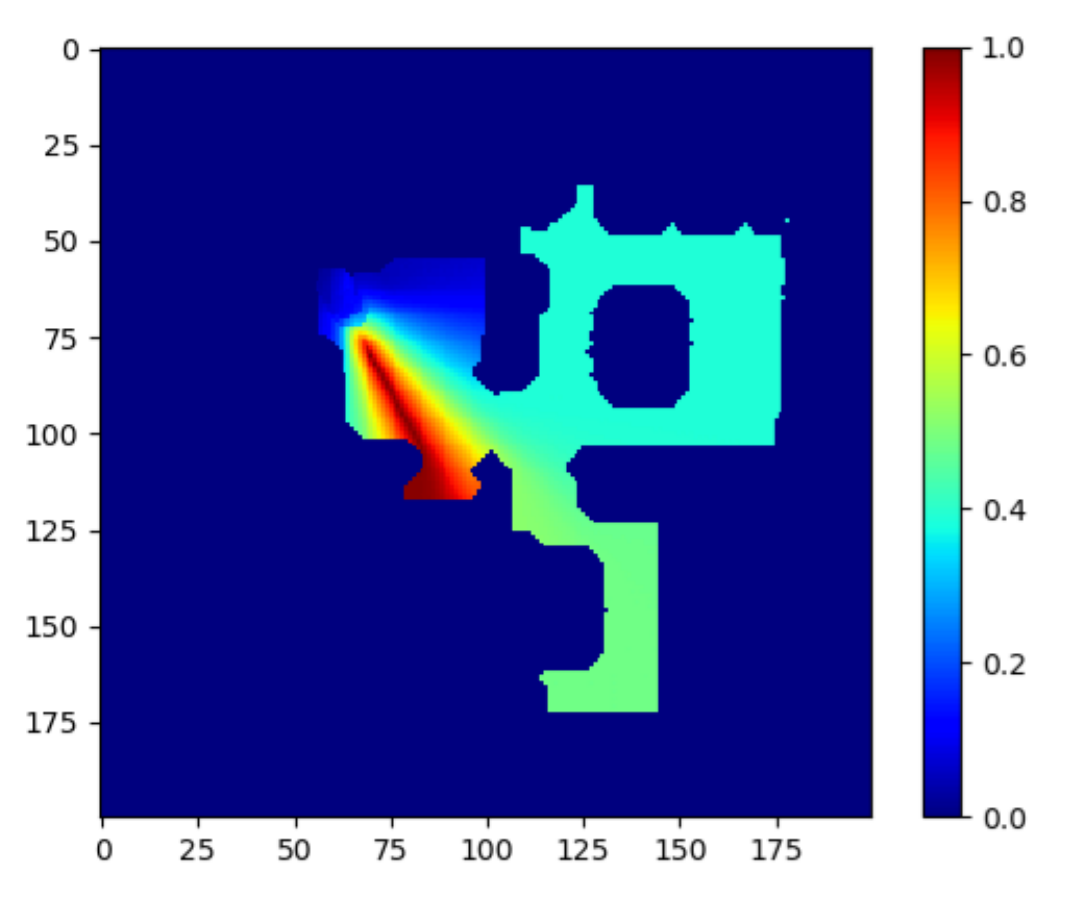}}
    \end{subfigmatrix}
    \caption{Data Preparation for Neural Network Training}
    \label{fig:training_data}
\end{figure}

This section describes how to create the dataset for training the model presented in section \ref{section:model_architecture}). 
The dataset consists of $M$, $\bm{x}_0$, $\bm{x}_t$, $\bm{v}^{user}_t$ and 
\begin{math} P(\bm{g} \mid \bm{v}^{user}_{t}, \bm{x}_t, M) \end{math}. 
The following outlines the procedure for creating each data:
\begin{enumerate} 
\renewcommand{\labelenumi}{(\arabic{enumi})}
    \item {Generating the map $M$}\\
    We acquire the map using SLAM (exemplified in Fig.\ref{fig:training_data}(a)). 
    For safe navigation, the map was modified using a heuristic method (shown in Fig.\ref{fig:training_data}(b)).
    
        

    \item {Generating 
    $\bm{x}_0$, $\bm{x}_t$, and $\bm{v}^{user}_t$
    }\\
    The robot's initial position $\bm{x}_0$, 
    self-position $\bm{x}_t$, 
    and user command $\bm{v}^{user}_t$ at time $t$, 
    are created through the following steps:
    \begin{enumerate}
    \renewcommand{\labelenumi}{(\arabic{enumi})}
        \item Setting the initial and goal positions\\
        In the map created in (1), pairs of points that are not on obstacles and whose Euclidean distance between them exceeds a certain threshold are adopted as the initial and goal positions.
        \item Generating robot's self-position $\bm{x}_t$ and user command $\bm{v}_t$\\
        Planning a route using the potential method for the initial and goal positions set in a).
        Saving points on the route as the robot's self-position at time $t$, 
        and calculating the robot's velocity $\bm{v}_t = (v_{x,t}, v_{y,t})$ using the following equations:
            \begin{align}
                v_{x,t} &= \frac{-\{ \phi(x_t+\Delta \mid \bm{g}, M) - \phi(x_t-\Delta \mid \bm{g}, M) \} \cdot s}{2 \cdot \Delta \cdot l}
                \label{eq:v_x}\\
                v_{y,t} &= \frac{-\{ \phi(y_t+\Delta \mid \bm{g}, M) - \phi(y_t-\Delta \mid \bm{g}, M) \} \cdot s}{2 \cdot \Delta \cdot l}
                \label{eq:v_y}
            \end{align}
        Here, $\phi$ represents the potential field when $\bm{g}$ is considered the goal, 
        $\bm{g}$ is the goal position set in a), 
        and $l$ is the magnitude of $\bm{v}_t$, calculated as
            \begin{equation}
                l = \sqrt{{v_{x,t}}^2 + {v_{y,t}}^2} .
                \notag
            \end{equation}
        $s$ is the robot's maximum speed in its travel direction, 
        and $\Delta$ is a sufficiently small constant.
        These values $v_{x,t}$ and $v_{y,t}$ are saved as user commands.
    \end{enumerate}
    
    \item {Generating \begin{math} P(\bm{g} \mid \bm{v}^{user}_{t}, \bm{x}_t, M) \end{math}}\\
    Using the robot's self-position $\bm{x}_t$ and user command $\bm{v}^{user}_{t}$ created in b), the probability distribution is calculated based on Equation (\ref{probability_t}).
    An example of the calculated probability distribution is shown in Fig. \ref{fig:training_data}(c).
\end{enumerate}

\subsubsection{Goal Selection}
Within the probability distribution expressed by Equation (\ref{probability_T}), 
the location with the highest probability value represents the user's intended goal. 
However, given the limitations on the input frequency and the discreteness of users' commands, 
there may be multiple points with similarly high probability values. 
Therefore, in this study, 
we determined the user's intended goal $\bm{g}_t$ as follows.
\begin{equation}
\label{eq:goal_selection}
    \bm{g}_t = \underset{\bm{g} \in G} {\operatorname{argmin}} \sqrt{{\bm{x}_t}^2 - {\bm{g}}^2}
\end{equation}
where, 
\begin{eqnarray}
    G=\{ \bm{g} \mid 
    P(\bm{g} \mid \bm{v}^{user}_{t-N:t}, \bm{x}_t, M)
    > T(p_{\max}-p_{\min}) \}, \nonumber \\
    \left \{
    \begin{array}{l}
    p_{\max} = \max P(\bm{g} \mid \bm{v}^{user}_{t-N:t}, \bm{x}_t, M)~~~~~~~~~ \\
    p_{\min} = \min P(\bm{g} \mid \bm{v}^{user}_{t-N:t}, \bm{x}_t, M)
    \end{array}
    \right . \nonumber
\end{eqnarray}
Here, $T$ represents the threshold for probability values, with the goal selected from among those that exceed this threshold. 
In this case, $T=0.95$.
\subsubsection{Confidence level of estimation}
The confidence level of the estimated goal $c(\bm{g}_t)$ is defined by the following formula:
\begin{gather}
\label{confidence_level}
    c(\bm{g}_t) = \beta
    \frac{P(\bm{g}_t \mid \bm{v}^{user}_{t-N:t}, \bm{x}_t, M)}{\sum_{\bm{g} \in G}  P(\bm{g} \mid \bm{v}^{user}_{t-N:t}, \bm{x}_t, M)}.
\end{gather}
Here, $\beta$ is a hyperparameter and set to 4.

\subsection{Method of Generating SC Commands}
This section explains the method employed by the shared controller to generate a SC command by combining the users and autonomous commands. 
Let the user command at time $t$ be $\bm{v}^{user}_{t}$, 
and the autonomous command calculated to head towards the estimated goal $\bm{g}_{t}$ be $\bm{v}^{auto}_{t}(\bm{g}_{t})$ (calculated using the Robot Autonomy Module), SC command $\bm{v}^{shared}_{t}$ is defined as follows: 
\begin{gather}
\label{a_sha}
    \bm{v}^{shared}_{t} = 
    c(\bm{g}_t) \cdot \bm{v}^{auto}_{t}(\bm{g}_t) + \{ 1-c(\bm{g}_t) \} \cdot \bm{v}^{user}_t 
\end{gather}
However, note that $t$ is continuous, and $\bm{v}^{user}_{t}$ holds the value entered in the previous step from the time it was input by the user until the next value was entered.
Ultimately, this SC command $\bm{v}^{shared}_{t}$ is input into the robot. 
Defining the SC command with this equation allows the system's autonomy to take precedence when the confidence in the estimation is high, and for the human user to primarily control the robot when it is low.
Furthermore, the SC command, which has a sufficiently high input frequency and is continuous compared to the user commands, can solve issues associated with the commands generated by the BMI.

\section{EXPERIMENT}
In this study, navigation experiments were conducted using a pseudo-BMI system.
The experiment consisted of two settings.
In Experiment 1, user commands were systematically generated to evaluate the navigation performance of the proposed SC system.
This experiment was conducted in a simulated environment.
In Experiment 2, participants used the proposed system to assess its performance with commands generated by humans, as well as to evaluate the usability of the proposed system.
This experiment was conducted in a real-world setting.
\subsection{Experimental Setup}
\subsubsection{Robot}
The experiments employed the Human Support Robot (HSR)\footnote{\url{https://global.toyota/jp/detail/8709536} }
developed by TOYOTA MOTOR CORPORATION as shown in Fig. \ref{fig:experimental_setup}(a). 
The HSR is capable of omnidirectional movement owing to its all-direction cart, and it can acquire information about its surroundings using a LiDAR at its base and an RGB-D camera on its head.
\subsubsection{Experimental Environment}
\begin{figure}[t]
    \begin{subfigmatrix}{2}
        \subfigure[robot]{\includegraphics[width=0.24\linewidth]{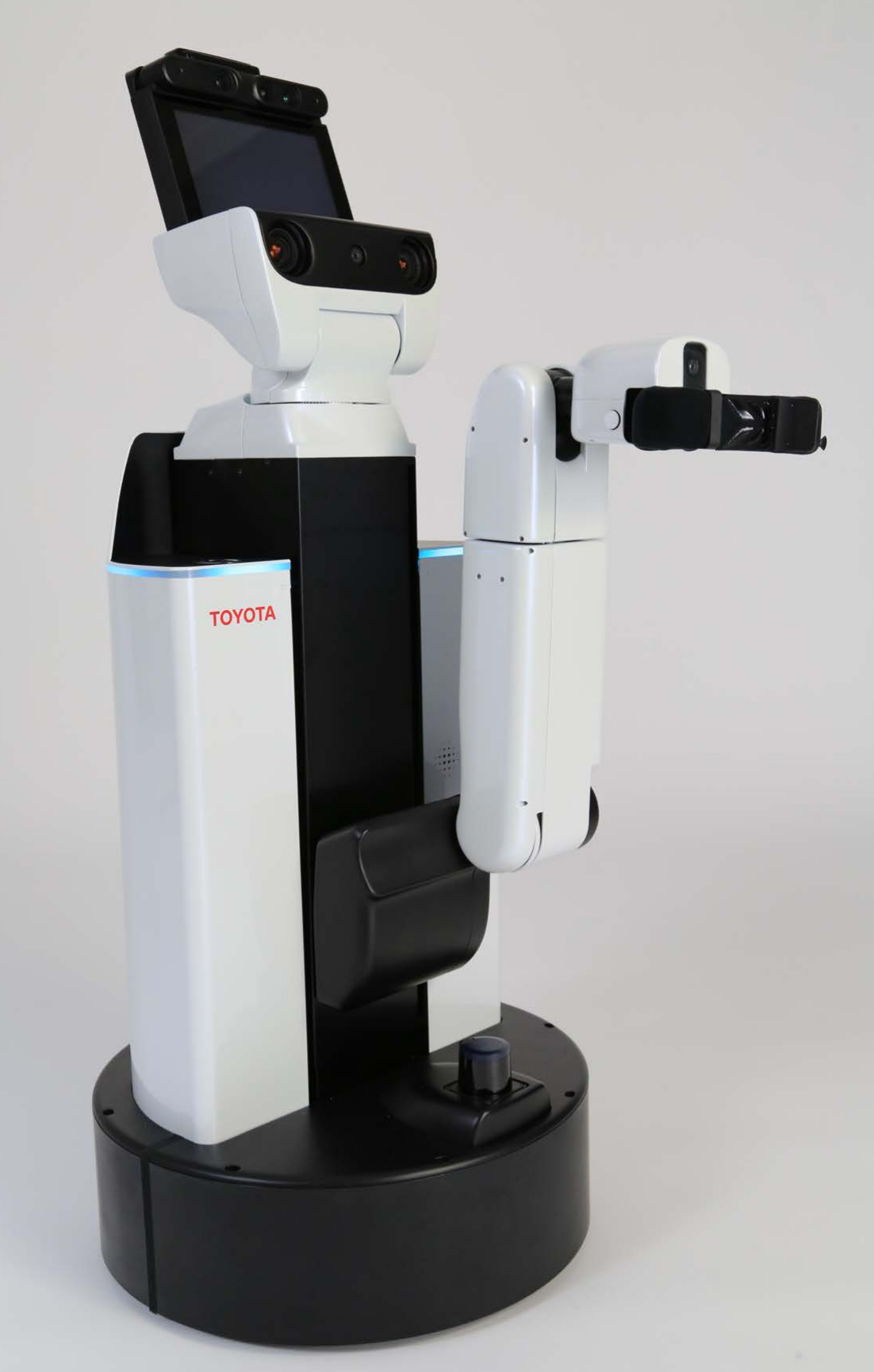}}
        \hspace*{-0.24cm}
        \subfigure[kitchen and living room]{\includegraphics[width=0.75\linewidth]{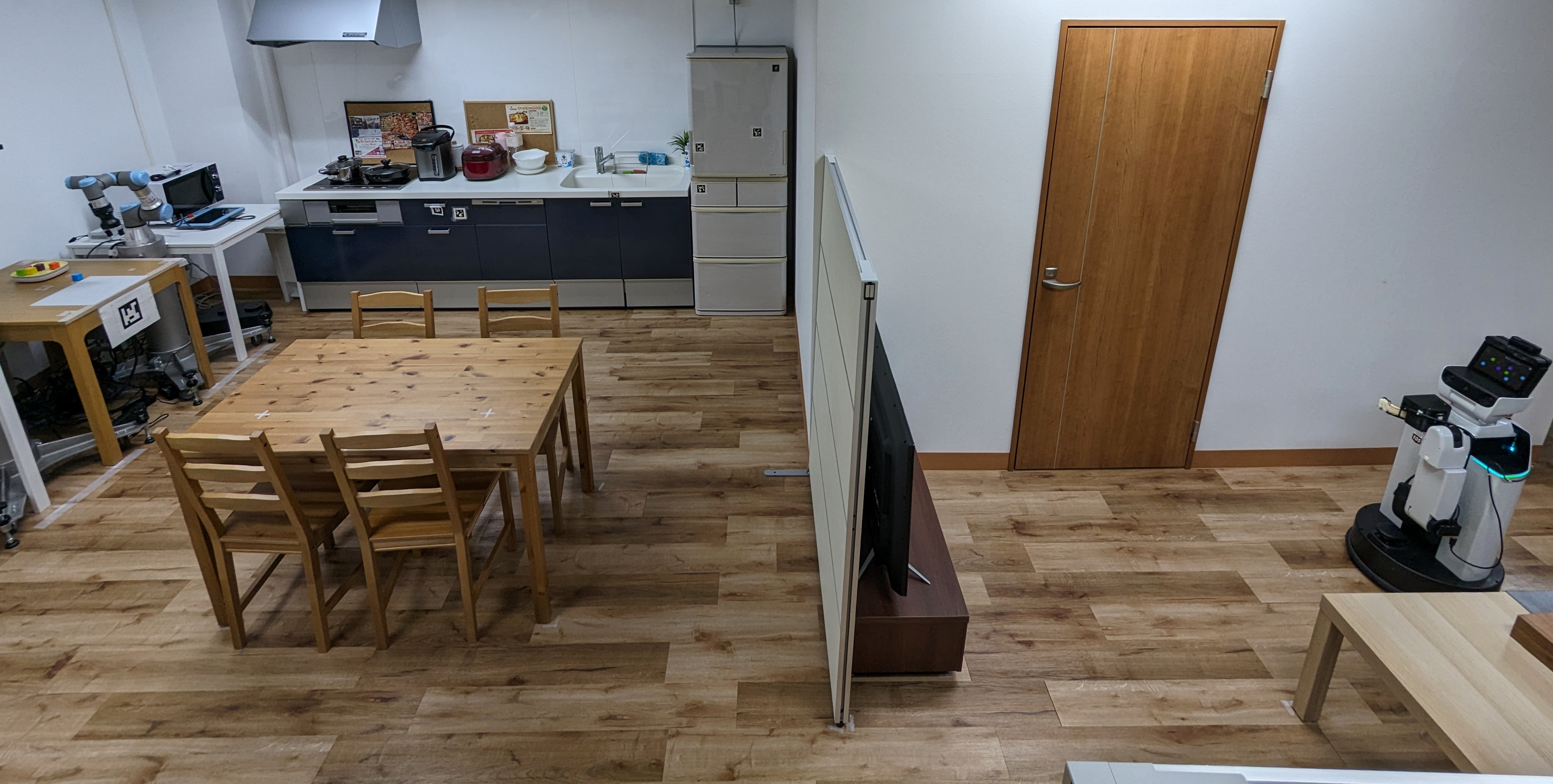}}
        \hspace*{0.8cm}
    \end{subfigmatrix}
    \caption{Experimental Setup}
    \label{fig:experimental_setup}
\end{figure}

The environment used for the experiments is depicted in Fig. \ref{fig:experimental_setup}(b), consisting of a kitchen and a living room, designed to simulate a domestic setting. 
The map provided to the robot in this environment is shown in Fig. \ref{fig:training_data}(b).

\subsubsection{Experimental Conditions}
\label{section:user_input_conditions}
In this experiment, users input the direction they wish to move the robot. Various conditions are applied to the user's input: 
\begin{enumerate} 
    \item {\bf Limitations on Input Frequency}\\
    The frequency at which the user can input commands is limited.
    In the experiment, it was set to 1 Hz.
    
    \item {\bf Limitations on Input Directions}\\
    The experiments compare three types of input directions: 
    all-directions, allowing unrestricted movement in any direction; 
    8-directions, which combine cardinal and diagonal directions for more nuanced control; 
    and 4-directions, limiting the robot to basic forward, backward, left, and right movements.
    In Experiment 2, we utilized corresponding UIs as shown in Fig. \ref{fig:user_input_devices}.
    
    \item {\bf Limitation on Input Accuracy}\\
    When the input directions were limited to four or eight, 
    some of the commands generated by the user were intentionally altered to incorrect commands with a certain probability and transmitted to the system. 
    This probability determines the input accuracy. 
    For example, at 100\% input accuracy, all the user inputs were correctly transmitted to the system. 
    At 90\% accuracy, 90\% of the inputs were correctly transmitted, whereas 10\% were forcibly changed to different inputs before being transmitted. 
    Experiments were conducted with input accuracies of 100, 90, 80, and 70\%.
    Input miscommunication occurs probabilistically according to these conditions, implying that the number and timing of miscommunications can vary even under the same input accuracy condition.

\end{enumerate}
\subsubsection{Task}
The experiment's task was to navigate the robot from an initial position in front of a living room table (right side of Fig.\ref{fig:experimental_setup}(b)) to a goal position in front of a refrigerator in the kitchen (top center of Fig.\ref{fig:experimental_setup}(b)), 
avoiding collisions with obstacles, and aiming for the fastest and shortest possible route.
\subsection{Experiment 1}
In Experiment 1, the robot was moved by automatically generating user commands according to certain calculation rules. 
The user command at time $t$, $\bm{v}^{user}_{t} = (v_{x,t}, v_{y,t})$ is generated using Equations (\ref{eq:v_x}) and (\ref{eq:v_y}). 
However, in the conditions where the input directions are limited to four or eight directions, 
the user command is defined by quantizing the calculated $\bm{v}^{user}_t$ into four or eight directions. 

In this experiment, the input accuracy varied between 100, 90, 80, and 70\%, and the task was performed under these conditions without applying SC.
Consequently, the task was executed under 18 different conditions ((one input direction + two input directions $\times$ four accuracy levels) $\times$ two control modes).
Each condition was tested five times. 

The evaluation metrics included the number of collisions with obstacles, the time required to complete the task, and the length of the route taken.

\subsection{Experiment 2}
In Experiment 2, ten participants (six males and four females, with an average age of 24.4 years and a standard deviation of 0.52 years) performed the navigation task. 

\begin{figure}[t]
    \begin{subfigmatrix}{3}
        \hspace*{0.1cm}
        \subfigure[joystick\newline (All-directions)]{\includegraphics[width=0.3\linewidth]{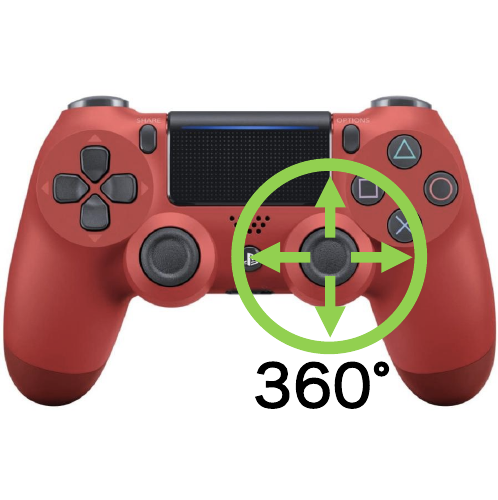}}
        \hspace*{-0.2cm}
        \subfigure[GUI (8-directions)]{\includegraphics[width=0.3\linewidth]{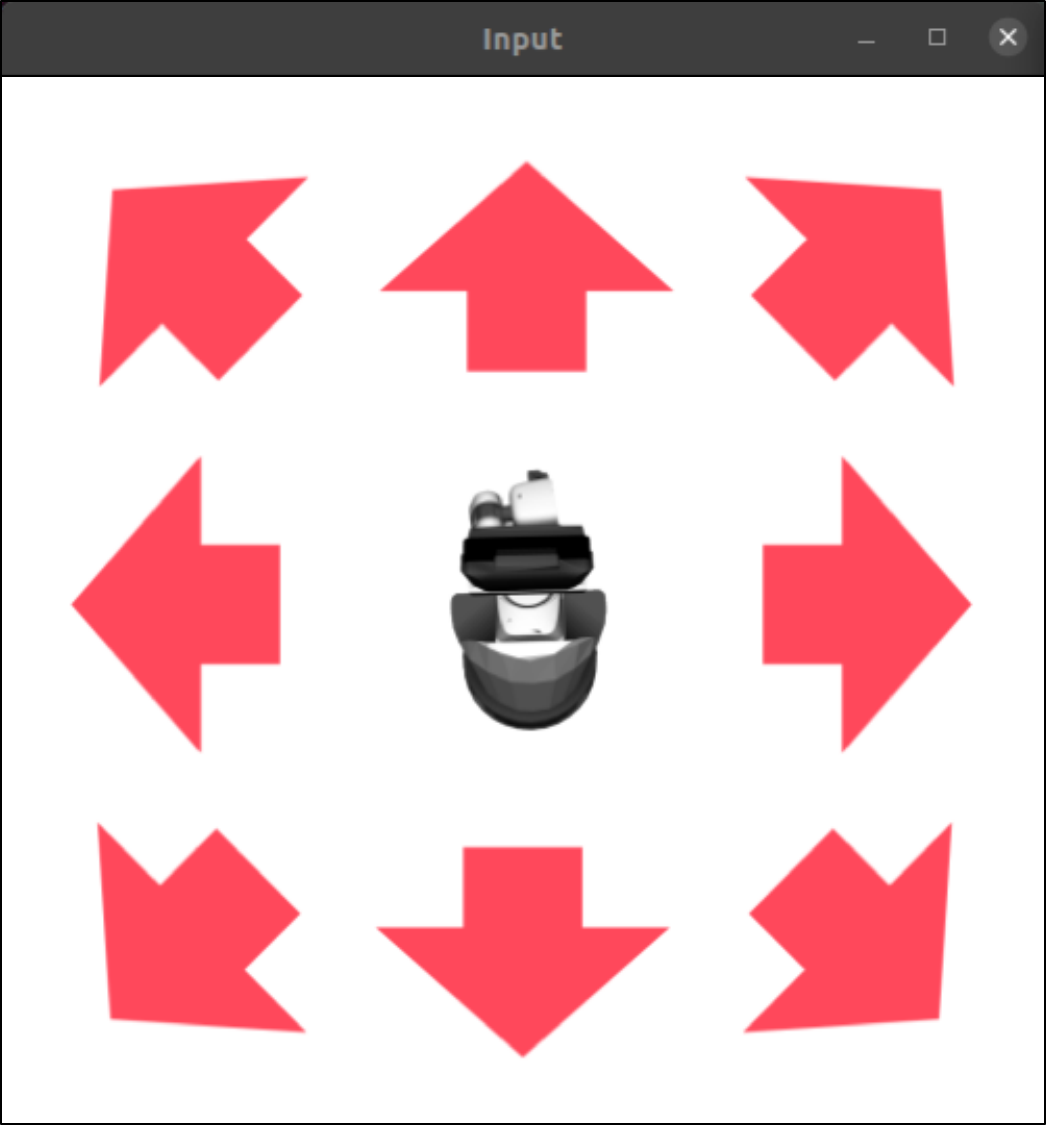}}
        \subfigure[GUI (4-directions)]{\includegraphics[width=0.3\linewidth]{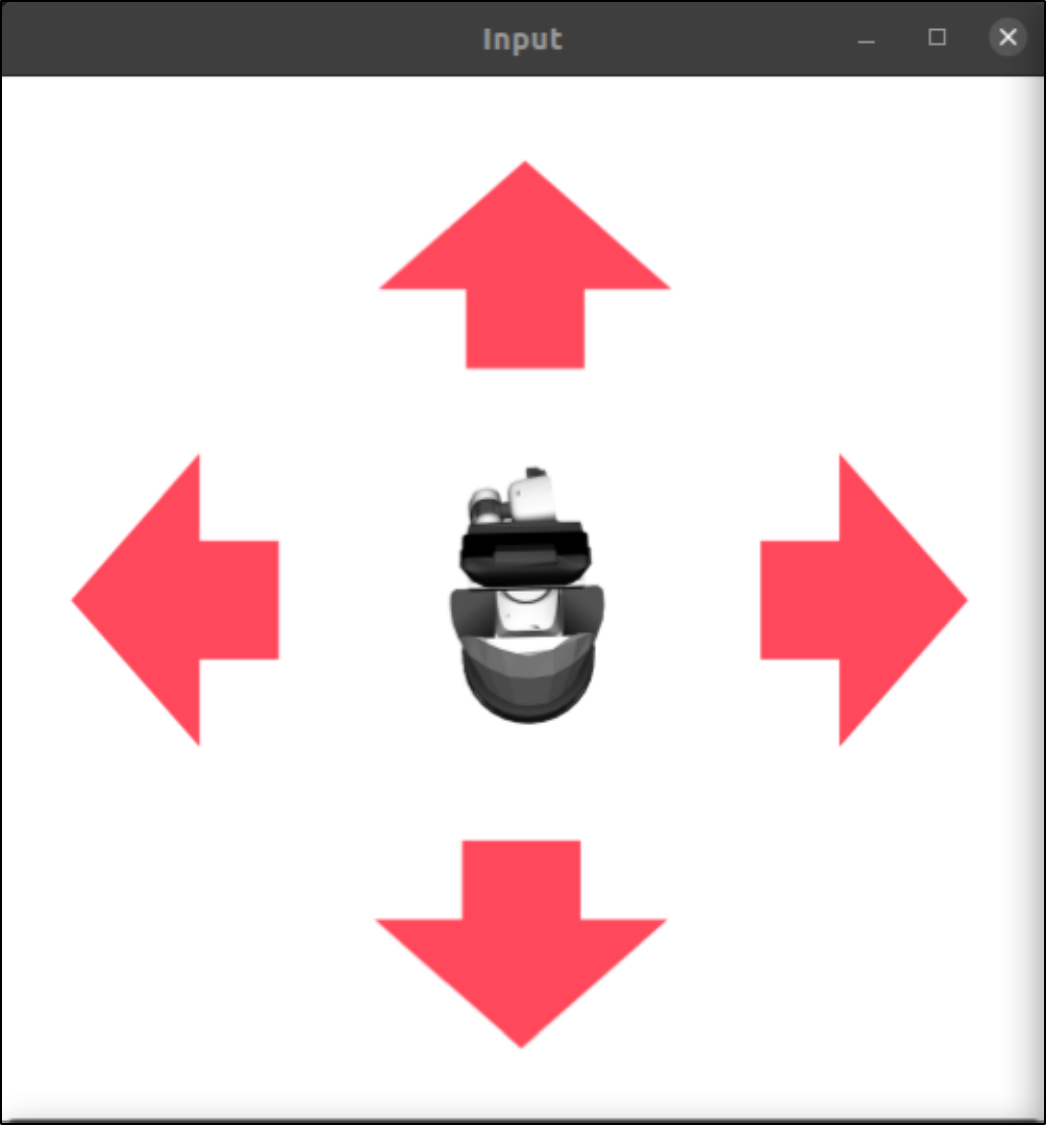}}
        \hspace*{-0.2cm}
    \end{subfigmatrix}
    \caption{The interfaces used to generate user commands}
    \label{fig:user_input_devices}
\end{figure}

The participants generated user commands using a joystick or GUIs, as shown in Fig.\ref{fig:user_input_devices}.
These devices were designed to apply the constraints described in Section \ref{section:user_input_conditions} to the user-generated commands.
To consider the burden on participants, the input accuracy was set to three levels: 100, 90, and 80\%. 
Consequently, the participants performed the task under 14 different conditions ((one input direction + two input directions $\times$ three accuracy levels) $\times$ two control modes).
The order of the conditions was randomly assigned and varied among the participants.
Before each trial began, the participants were informed about the conditions (input frequency, number of directions, input accuracy, and whether SC is applied).

The evaluation metrics included 
the success rate of the task (defined as reaching the goal position without colliding with obstacles), required time, and path length.
However, the time and path length were measured only when the task was successfully completed.
In addition, the participants answered a questionnaire at the end of each trial. 
The questionnaire consisted of the following items.
\begin{enumerate}
    \item Was the robot easy to operate?
    \item Did you feel confident operating the robot?
    \item Was the robot's movement speed comfortable?
    \item Did the robot move according to your instructions? 
\end{enumerate}
Participants rated the questions on a seven-point scale (0=disagree to 6=agree).
We analyzed the survey scores as a Likert scale.

\section{Results}
\subsection{Experiment 1}
\begin{table*}
\centering
\caption{Evaluation of Experiment 1 and 2: 
“Collisions” counts robot-obstacle collisions, 
"Times" is task completion time, 
"Path Length" measures the route length from start to target, 
and "Success Rate" is the percentage of reaching the target without collisions. 
Underlines indicate statistical significance: 
\dashuline{dashed} (p$<$0.05), 
\underline{single} (p$<$0.01), 
and \doubleunderline{double lines} (p$<$0.001).
}
\label{tab:experimental_result}
\scriptsize
\begin{tabular}{|ll||cc|cc|cc||cc|cc|cc|}
\hline
&& \multicolumn{6}{c||}{EXPERIMENT I} & \multicolumn{6}{c|}{EXPERIMENT II}\\
\hline
&& \multicolumn{2}{c|}{Collisions} & \multicolumn{2}{c|}{Times [s]} & \multicolumn{2}{c||}{Path length [m]} & 
\multicolumn{2}{c|}{Success rate [\%]} & \multicolumn{2}{c|}{Times [s]} & \multicolumn{2}{c|}{Path length [m]} \\
\multicolumn{2}{|c||}{Cond.}& w SC & w/o SC &  w SC & w/o SC & w SC &  w/o SC
& w SC & w/o SC &  w SC & w/o SC & w SC &  w/o SC\\
\hline
\hline
All  & \hspace*{-0.35cm} 100 & 
0.0 $\pm$ 0.0 \hspace*{-0.35cm} & 0.0 $\pm$ 0.0 & 
24.4 $\pm$ 0.3 \hspace*{-0.35cm} & 21.7 $\pm$ 0.4 &
\dashuline{$\bm{5.1 \pm 0.1}$} \hspace*{-0.35cm} & 5.3 $\pm$ 0.1 &
100 \hspace*{-0.35cm} & 100 & 
26.1 $\pm$ 1.2 \hspace*{-0.35cm} & \underline{23.8 $\pm$ 1.4} &
6.1 $\pm$ 0.3 \hspace*{-0.35cm} & 6.1 $\pm$ 0.3
\\
\hline
8 & \hspace*{-0.35cm} 100 &
0.0 $\pm$ 0.0 \hspace*{-0.35cm} & 0.0 $\pm$ 0.0 & 
24.9 $\pm$ 0.2 \hspace*{-0.35cm} & \dashuline{23.2 $\pm$ 0.6} &
\underline{$\bm{5.3 \pm 0.0}$} \hspace*{-0.35cm} & 5.5 $\pm$ 0.1 &
100 \hspace*{-0.35cm} & 100 &
6.4 $\pm$ 0.5 \hspace*{-0.35cm} & 5.9 $\pm$ 0.2 &
28.7 $\pm$ 3.0 \hspace*{-0.35cm} & \underline{23.2 $\pm$ 1.3} 
\\
8 & \hspace*{-0.35cm} 90 &
0.2 $\pm$ 0.4 \hspace*{-0.35cm} & 0.6 $\pm$ 0.5 & 
27.5 $\pm$ 1.3 \hspace*{-0.35cm} & 25.7 $\pm$ 2.1 &
5.4 $\pm$ 0.1 \hspace*{-0.35cm} & 6.0 $\pm$ 0.4 &
$\bm{90}$ \hspace*{-0.35cm} & 60  &
33.7 $\pm$ 11.3 \hspace*{-0.35cm} & 25.9 $\pm$ 12.7 &
7.0 $\pm$ 2.2 \hspace*{-0.35cm} & 6.6 $\pm$ 3.2
\\
8 & \hspace*{-0.35cm} 80 &
\dashuline{$\bm{0.0 \pm 0.0}$} \hspace*{-0.35cm}  & 4.2 $\pm$ 2.1 & 
31.8 $\pm$ 3.5 \hspace*{-0.35cm}  & 32.4 $\pm$ 2.3 &
\dashuline{$\bm{5.7 \pm 0.3}$} \hspace*{-0.35cm}  & 7.2 $\pm$ 0.6 &
$\bm{100}$ \hspace*{-0.35cm} & 70  & 
37.3 $\pm$ 7.4 \hspace*{-0.35cm} & 34.3 $\pm$ 16.2 &
7.5 $\pm$ 0.9 \hspace*{-0.35cm} & 8.7 $\pm$ 4.1
\\
8 & \hspace*{-0.35cm} 70 &
\underline{$\bm{0.0 \pm 0.0}$} \hspace*{-0.35cm} & 4.4 $\pm$ 1.6 & 
30.7 $\pm$ 1.0 \hspace*{-0.35cm} & 38.3 $\pm$ 6.9 &
\dashuline{$\bm{5.6 \pm 0.1}$} \hspace*{-0.35cm} & 8.3 $\pm$ 1.3 &
- \hspace*{-0.35cm} & - & 
- \hspace*{-0.35cm} & - & 
- \hspace*{-0.35cm} & -
\\
\hline
4 & \hspace*{-0.35cm} 100 &
\doubleunderline{$\bm{0.0 \pm 0.0}$} \hspace*{-0.35cm} & 1.8 $\pm$ 0.4 & 
26.0 $\pm$ 0.3 \hspace*{-0.35cm} & 27.2 $\pm$ 1.2 &
\doubleunderline{$\bm{5.3 \pm 0.1}$} \hspace*{-0.35cm} & 6.0 $\pm$ 0.1 &
100 \hspace*{-0.35cm} & 100 & 
32.3 $\pm$ 2.1 \hspace*{-0.35cm} & \underline{25.6 $\pm$ 1.8} &
6.8 $\pm$ 0.3 \hspace*{-0.35cm} & 6.4 $\pm$ 0.3
\\
4 & \hspace*{-0.35cm} 90 &
\dashuline{$\bm{0.0 \pm 0.0}$} \hspace*{-0.35cm} & 2.2 $\pm$ 1.2 & 
30.0 $\pm$ 2.4 \hspace*{-0.35cm} & 29.4 $\pm$ 2.1 &
\underline{$\bm{5.6 \pm 0.2}$} \hspace*{-0.35cm} & 6.5 $\pm$ 0.4 &
$\bm{100}$ \hspace*{-0.35cm} & 70  & 
43.1 $\pm$ 8.0 \hspace*{-0.35cm} & 30.6 $\pm$ 14.6 &
7.7 $\pm$ 1.0 \hspace*{-0.35cm} & 7.4 $\pm$ 3.5
\\
4 & \hspace*{-0.35cm} 80 &
\dashuline{$\bm{0.0 \pm 0.0}$} \hspace*{-0.35cm} & 3.8 $\pm$ 2.4 &
31.1 $\pm$ 3.8 \hspace*{-0.35cm} & 38.6 $\pm$ 9.1 &
\dashuline{$\bm{5.6 \pm 0.4}$} \hspace*{-0.35cm} & 8.2 $\pm$ 1.8 &
$\bm{90}$ \hspace*{-0.35cm} & 20 &
45.1 $\pm$ 16.2 \hspace*{-0.35cm} & 42.8 $\pm$ 17.2 &
\dashuline{8.0 $\pm$ 2.5} \hspace*{-0.35cm} & 10.1 $\pm$ 4.1
\\
4 & \hspace*{-0.35cm} 70 &
\dashuline{$\bm{0.0 \pm 0.0}$} \hspace*{-0.35cm} & 6.8 $\pm$ 3.7 &
\dashuline{$\bm{34.1 \pm 4.0}$} \hspace*{-0.35cm} & 52.1 $\pm$ 7.2 &
\underline{$\bm{5.8 \pm 0.4}$} \hspace*{-0.35cm} & 10.8 $\pm$ 1.3 &
- \hspace*{-0.35cm} & - & 
- \hspace*{-0.35cm} & - & 
- \hspace*{-0.35cm} & -
\\
\hline
\end{tabular}
\normalsize 
\end{table*}

\subsubsection{Number of Collisions with Obstacles}
\label{section:ex1_collision}
The average number of collisions between the robot and obstacles under each condition is listed in TABLE \ref{tab:experimental_result}. 
Under conditions without SC, 
the number of collisions tends to increase as the input accuracy decreases. 
By contrast, under conditions with SC, the number of collisions does not increase even if the input accuracy decreases and remains at a constant level. 
\subsubsection{Task Completion Times}
\label{section:ex1_time} 
As shown in TABLE \ref{tab:experimental_result}, 
when the input accuracy is sufficiently high and the number of input directions is sufficiently large, 
executing the task without SC results in a shorter required time than when SC is applied. 
This can be attributed to the fact that under conditions without SC, the robot operates at its maximum speed in the intended direction, 
whereas in the case of SC, the absolute value of the velocity vector is reduced owing to the summation of the different velocity vectors. 
However, under the condition of four input directions and 70\% input accuracy, the required time is significantly shorter with SC. 
Although not statistically significant, a similar trend is observed under the condition with eight input directions and 70\% input accuracy, and four input directions with 80\% input accuracy.
Therefore, it can be inferred that applying SC can maintain the required time at a consistent level, despite a decrease in the number of input directions or input accuracy. 
\subsubsection{Path Length}
\label{section:ex1_path_length}
\begin{figure}[t]
    \begin{subfigmatrix}{2}
        \subfigure[8 directions]{\includegraphics[width=0.49\linewidth]{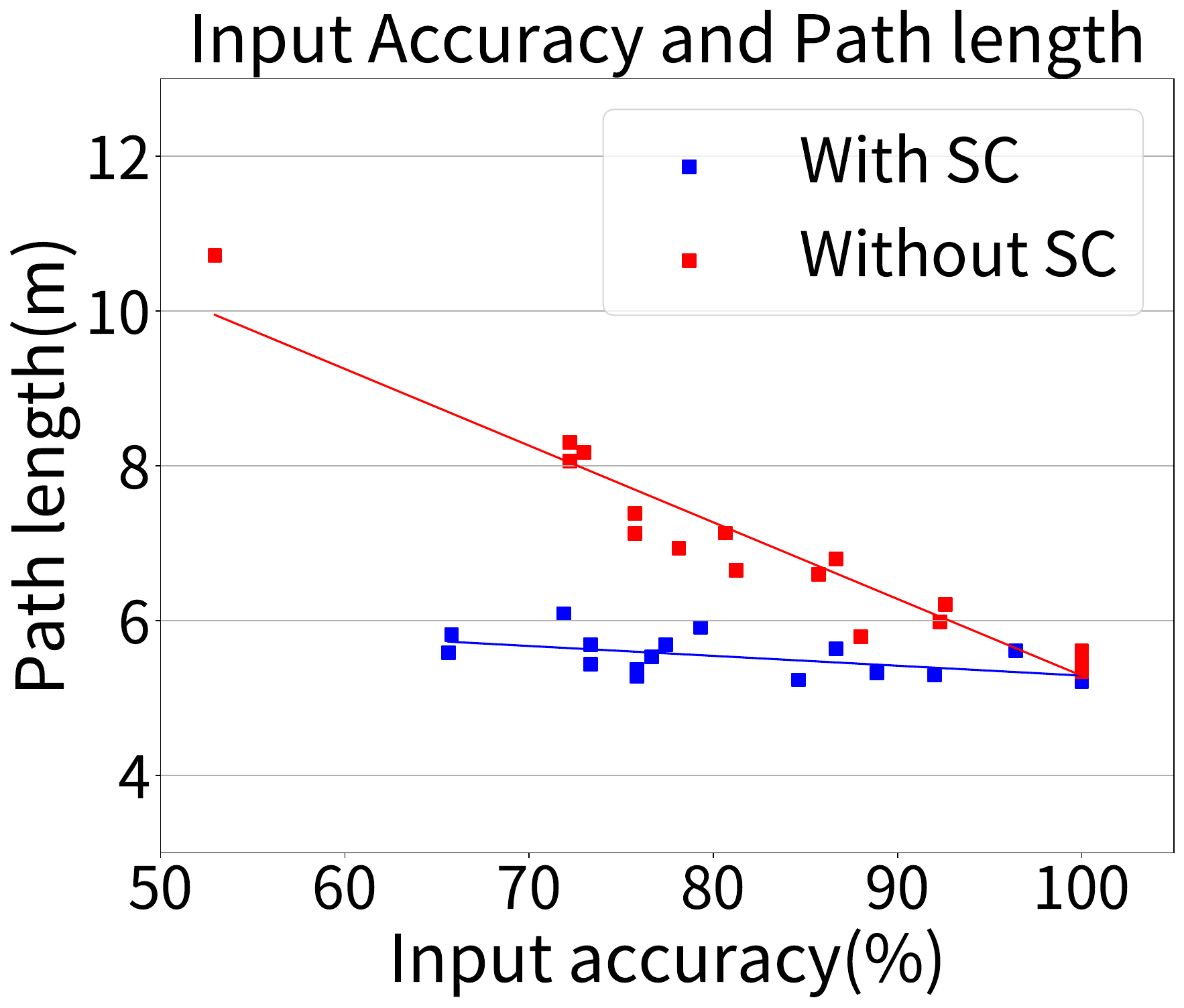}}
        \hspace*{-0.15cm}
        \subfigure[4 directions]{\includegraphics[width=0.49\linewidth]{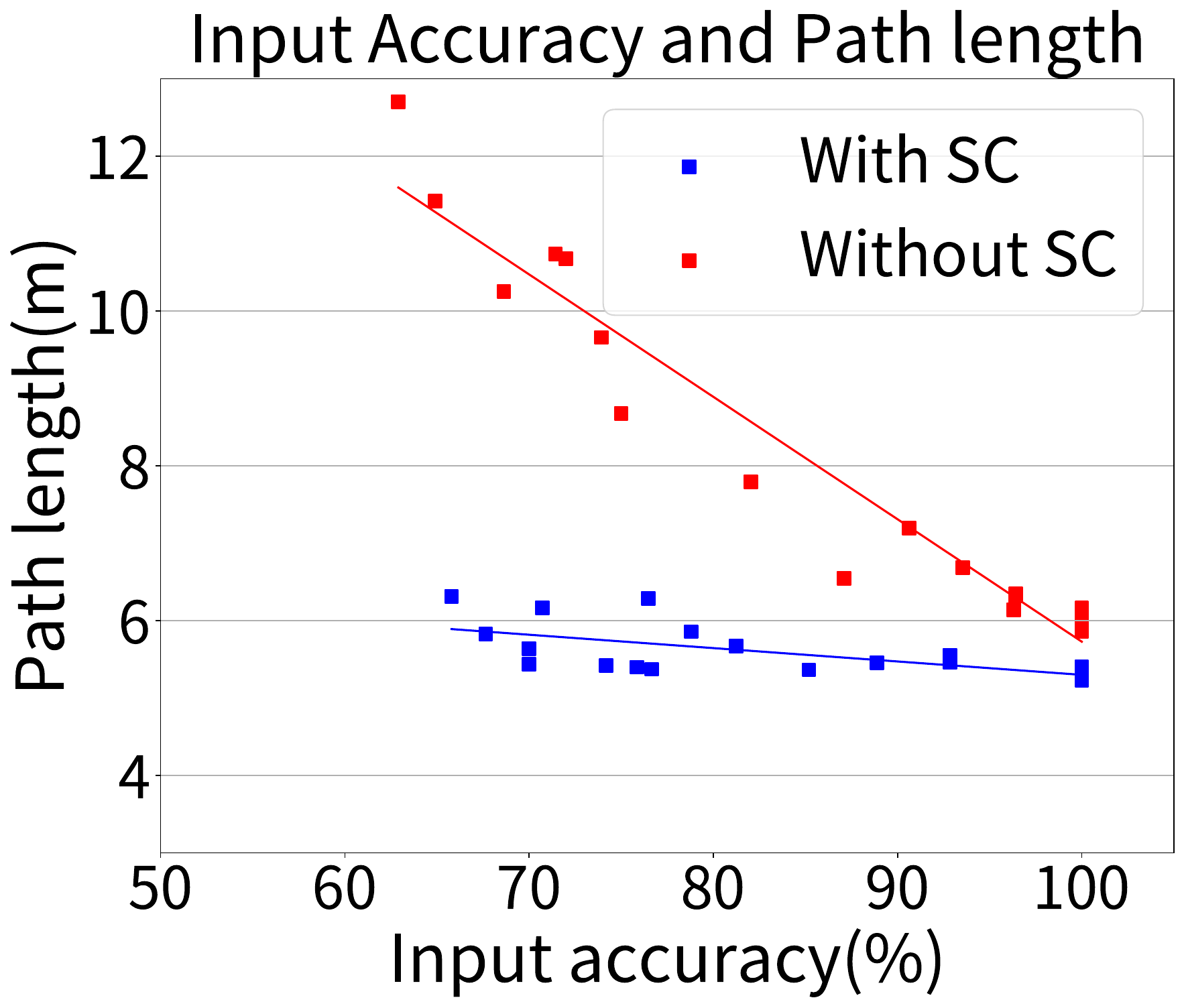}}
    \end{subfigmatrix}
    \caption{Relationship Between Input Accuracy and Path Length}
    \label{fig:sim_inputacc_path_length}
\end{figure}

The application of SC significantly reduces the path length compared to conditions without SC, as shown in TABLE \ref{tab:experimental_result}. 
Without SC, path length significantly increases as input accuracy decreases. 
However, with SC, it is possible to achieve a path length comparable to that of high-accuracy scenarios even with declining input accuracy.
Additionally, the relationship between the input accuracy and path length for both eight and four input directions is illustrated in Fig.\ref{fig:sim_inputacc_path_length}, where the horizontal axis represents the input accuracy and vertical axis denotes the path length. 
The regression line indicates that input accuracy significantly affects navigation performance, particularly when the number of controllable directions is reduced.
SC effectively maintained the performance despite reduced accuracy, in contrast to the noticeable decline in performance without SC. 
This is particularly relevant because the operation of BMI-driven robots often limits the number of controllable directions to approximately four. 

\subsection{Experiment 2}
\subsubsection{Success Rate}
\label{section:ex2_success_rate}
The success rates for each condition of the navigation tasks are listed in TABLE \ref{tab:experimental_result}.
This shows that the application of SC enables reliable navigation to the goal without obstacle collisions, even when the input accuracy is compromised.

\subsubsection{Task Completion Times}
\label{section:ex2_time}
From TABLE \ref{tab:experimental_result}, under conditions with 'All' and '8' input directions at 100\% accuracy, 
without SC statistically significantly reduced task completion time. 
This is consistent with the reasons mentioned in Section \ref{section:ex1_time}.
This trend was more pronounced in Experiment 2 than in Experiment 1, as discussed in Section \ref{section:discussion}.
\subsubsection{Path Length}
\label{section:ex2_path_length}
Under conditions of 8 input directions and 100\% input accuracy, not using SC resulted in significantly shorter path lengths.
Conversely, under the conditions of 4 input directions and 80\% input accuracy, the scenario with SC achieved significantly shorter path lengths.
This condition precisely reflects those encountered in BMI control. 
Our method worked well in the case like BMI's discrete and imprecise inputs.
\subsubsection{User Survey Results}
\begin{figure*}[t]
    \centering
    \includegraphics[keepaspectratio, scale=0.528]{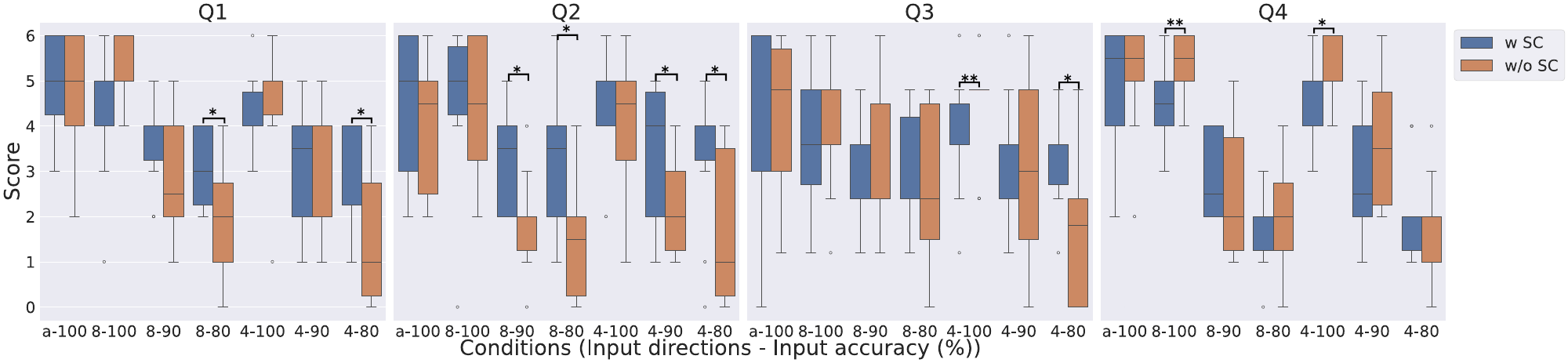}
    \caption{Scores for Q. 1-4: We treat the scores as based on a Likert scale. Asterisks (*) indicate statistical significance, with * p $<$ 0.05, ** p $<$ 0.01.}
    \label{fig:q_1_4}
\end{figure*}

The distribution of scores for Questions 1--4 across each condition is shown in Fig.\ref{fig:q_1_4}.
We considered the scores based on a Likert scale and conducted t-tests on the mean values.

The results for Q.1 indicate that under the condition of 80\% input accuracy, for both 4 and 8 input directions, scores were statistically significantly higher with SC conditions. 
This suggests that users feel that the operation is easier with SC when the input accuracy decreases.
The results for Q.2 show that under conditions of 90\% and 80\% input accuracy, for both 4 and 8 input directions, scores were statistically significantly higher with the SC conditions. 
This indicates that under conditions with input uncertainty, applying SC provides users with a sense of assurance during operation.
Furthermore, the results for Q.3 show that scores were statistically significantly higher without SC under conditions with 4 input directions and 100\% input accuracy. 
Conversely, under conditions with four input directions and 80\% input accuracy, scores were statistically significantly higher with SC. 
This suggests that SC is effective under the challenging conditions often encountered with BMI.
Finally, the results for Q.4 indicate that under conditions with 100\% input accuracy, scores were higher without SC. 
This is reasonable because SC can introduce slight discrepancies between the user commands and the robot's actual behavior, thus indicating the correct application of SC. 
Nevertheless, under some conditions, the scores were higher with SC, suggesting instances in which SC effectively mitigated the impact of command uncertainty.
\section{DISCUSSION}
\label{section:discussion}
\subsection{Effectiveness}
\subsubsection{Reduction in Collision}
As illustrated in Sections \ref{section:ex1_collision} and \ref{section:ex2_success_rate}, our proposed SC method significantly reduced collisions with obstacles in navigation tasks, 
enabling safer navigation.
\subsubsection{Shortening of Path Lengths}
From Section \ref{section:ex1_path_length}, Experiment 1, demonstrated that our method significantly shortened path lengths under almost all conditions.
%
%
Furthermore, as the number of controllable directions decreased, the performance deteriorated significantly under conditions without SC when the accuracy decreased. 
However, with SC, performance was maintained, allowing navigation with path lengths comparable to those under optimal conditions, which is a possible advantage when considering operations via BMI.

In Experiment 2, under the conditions of eight input directions and 100\% input accuracy, the paths were shorter without SC.
Under these conditions, the success rates were 100\% both with and without SC, indicating that the tasks could be easily accomplished.
Nevertheless, operating robots via BMI often presents more challenging scenarios, in which SC assistance becomes crucial. 
For example, in more challenging situations (e.g., with four input directions and 80\% input accuracy), the use of SC significantly shortened the path length. 
These are the conditions faced when considering BMI control, making this method particularly effective as inputs become more discrete and accuracy decreases, which are common characteristics of BMI inputs.
\subsubsection{Usability}
Survey responses to Questions 1, 2, and 3 reveal that 
with diminished input accuracy, users reported easier, more confident, and comfortably paced operation with SC. 
These findings suggest a favorable user perception of SC under conditions commonly encountered in BMI control.
\subsection{Limitations}
\label{limitations}
\subsubsection{Challenges in Collision Avoidance}
\label{limitations_1}
In both experiments, there were instances where collisions could not be avoided immediately after task start, even under conditions with SC.
Analysis revealed that, especially at the start, the low kurtosis of the probability distribution calculated by Equation (\ref{probability_T}) led to a low confidence level of the estimation derived from Equation (\ref{confidence_level}). 
Consequently, SC commands generated based on Equation (\ref{a_sha}) were heavily influenced by user commands, with minimal reflection of autonomous commands, making it difficult to compensate for the uncertainty in user commands. 
Integrating environmental information, such as proximity to obstacles, into the weighting could potentially mitigate these issues.

\subsubsection{Challenges in Increased Task Completion Time}
\label{limitations_2}
Implementing our SC method showed a potential increase in task completion time, particularly in Experiment 2.
As discussed in Sections \ref{section:ex1_time} and \ref{section:ex2_time}, whereas the robot moves at its maximum speed in a specific direction without SC, with SC, it moves according to the velocity vector calculated by Equation \ref{a_sha}, 
blending the user and autonomous commands that do not always align perfectly. 
This combination results in a slower velocity command than when the robot is operating at its maximum speed alone, thereby extending thecompletion times. 

Additionally, the survey results from Q.4 indicate that in SC, participants felt that the robot did not directly respond to their commands. 
Further interviews revealed that participants often mistook discrepancies between their intended commands and the robot's actual behavior for input recognition errors, leading users to input velocity commands in the opposite direction of the robot's movement (corrective movement commands). 
As a result, the velocity command ($\bm{v}^{shared}_{t}$) issued to the robot may have been further reduced. 
However, this issue may diminish as users gain familiarity with the SC system through repeated interactions.
This suggests that further validation of the experimental results after several practice sessions could provide deeper insights into the effectiveness of the SC method effectiveness.
\section{CONCLUSION}
In this study we introduced a SC method for BMI-based remote robot navigation, addressing challenges, such as low input frequency, discreteness, and noise-induced uncertainty in BMI-generated commands. 
We showed that our approach, which generates auxiliary commands by estimating the user's intended goal and blending these with user commands based on the confidence level of the estimation, significantly reduces obstacle collisions and shortens paths, particularly with more discrete and noisy user inputs.

Future directions will focus on improving the strategy for combining user and autonomous commands.
Our current approach, which uses the confidence level as the weight (defined in Equations (\ref{a_sha}) and (\ref{confidence_level})), leads to collisions right after starting and a decrease in the robot's speed, as shown in Sections \ref{limitations}. 
Therefore, we plan to integrate environmental factors, such as obstacle proximity and the reliability of user input, into the weighting.
Additionally, as an extra solution to the limitation outlined in Section \ref{limitations_1}, we plan to introduce a prior probability distribution. 
Our current method calculates the user's goal probability from user commands and environmental data, such as maps (Equation (\ref{probability_T})). 
For instance, indoor navigation often features areas with inherently higher or lower goal probabilities—flat floors, for instance, are less likely targets than areas near objects. 
Integrating a prior distribution based on these insights will help in estimating goals more quickly and reliably.

\addtolength{\textheight}{-12cm} 

\bibliographystyle{templete/IEEEtran} 
\bibliography{templete/IEEEabrv.bib, root.bib} 

\end{document}